\documentclass[letterpaper, 10 pt, conference]{ieeeconf}  

\IEEEoverridecommandlockouts                   
\usepackage{graphicx} 
\usepackage{amsmath}
\usepackage{amssymb}  
\usepackage{amsfonts}
\usepackage{algorithmic}
\usepackage{array}
\usepackage{textcomp}
\usepackage{stfloats}
\usepackage{url}
\usepackage{verbatim}
\usepackage{graphicx}
\usepackage{caption}
\usepackage{subcaption}
\usepackage[ruled, vlined, linesnumbered]{algorithm2e}
\usepackage{cite}
\usepackage{amssymb,amsfonts,bm}
\usepackage{amsmath,tikz}
\usepackage{color}
\usepackage{mathtools}
\usepackage[hidelinks]{hyperref} 
\usepackage{rotating}
\usepackage{blkarray}
\usepackage{physics}
\usetikzlibrary{arrows}

\usepackage{amsthm}
\usepackage{comment}

\newcommand*{\tor}{\tau}
\newcommand*{\mass}{m}
\newcommand*{\pos}{p}
\newcommand*{\posvec}{\vb{p}}
\newcommand*{\tilt}{\sigma}
\newcommand*{\thrust}{T}
\newcommand*{\for}{f}

\newcommand*{\gbf}[1]{\boldsymbol{#1}}
\newcommand*{\distvec}{\vb{d}}
\newcommand*{\myset}[1]{\mathcal{#1}}
\newcommand*{\body}{\myset{B}}
\newcommand*{\inertial}{\myset{I}}
\newcommand*{\roll}{\phi}
\newcommand*{\pitch}{\theta}
\newcommand*{\yaw}{\gamma}
\newcommand*{\angvel}{\omega}
\newcommand*{\angvelvec}{\gbf{\omega}}
\newcommand*{\rotmatrix}{\vb{R}}

\newcommand*{\forcevec}{\vb{\for}}
\newcommand*{\torvec}{\gbf{\tor}}
\newcommand*{\screw}[1]{[#1]_\times}
\newcommand*{\angwheel}{\delta}
\newcommand*{\radius}{r}

\newcommand*{\moi}{\vb{J}}
\newcommand*{\bodywheel}{{\myset{W}}}
\newcommand*{\bodywheelone}{{\myset{W}_{1}}}
\newcommand*{\bodywheeltwo}{{\myset{W}_{2}}}
\newcommand*{\moicombined}{\vb{J}_{\Sigma}}
\newcommand*{\moiscaler}{J}
\newcommand*{\dist}{l}
\newcommand*{\friction}{F}

\newcommand*{\thrusterror}{T_e}

\newcommand*{\angvelvecdesir}{\angvelvec_d}
\newcommand*{\velcontrol}{\torvec}
\newcommand*{\wheelvel}{v}
\newcommand*{\wheelveldesired}{\wheelvel_d}
\newcommand*{\steerrate}{\dot{\yaw}}
\newcommand*{\airk}{\vb{K}^\omega}
\newcommand*{\airka}{\vb{K}^R_p}
\newcommand*{\airkp}{\airk_p}
\newcommand*{\airki}{\airk_i}
\newcommand*{\airkd}{\airk_d}

\newcommand*{\balancek}{K^b}
\newcommand*{\balancekv}{\balancek_v}
\newcommand*{\balancekp}{\balancek_p}
\newcommand*{\balanceki}{\balancek_i}
\newcommand*{\balancekd}{\balancek_d} 
\newcommand*{\balanceks}{\balancek_\yaw}
\newcommand*{\thr}{T_h}
\newcommand*{\thrin}{\thr^d}
\newcommand*{\thrhover}{\thr^{hover}}
\newcommand*{\thridle}{\thr^{idle}}

\newcommand*{\decouplek}{K^d} 
\newcommand*{\decoupleks}{\decouplek_s}
\newcommand*{\decouplekv}{\decouplek_v}
\newcommand*{\decouplekT}{K^t}
\newcommand*{\decouplekTp}{\decouplekT_p}
\newcommand*{\decouplekTi}{\decouplekT_i}
\newcommand*{\decouplekTd}{\decouplekT_d}
\newcommand*{\decoupleksigma}{K^\sigma}
\newcommand*{\decoupleksigmap}{\decoupleksigma_p}
\newcommand*{\decoupleksigmai}{\decoupleksigma_i}
\newcommand*{\decoupleksigmad}{\decoupleksigma_d}

\newcommand*{\uncouplethetae}{\angvel_{\pitch e}}
\newcommand*{\uncouplethetaedot}{\dot{\angvel}_{\pitch e}}
\newcommand*{\normal}{N}
\newcommand*{\horiz}{H}
\renewcommand{\vb}[1]{\boldsymbol{#1}}
\newcommand*{\scale}{C}
\newcommand*{\desiredsteer}{\angvel_{\yaw d}}
\newcommand*{\desiredpitchrate}{\angvel_{\pitch d}}
\newcommand*{\degree}{^{\circ}}

\title{\LARGE \bf
DoubleBee: A Hybrid Aerial-Ground Robot with Two Active Wheels
}

\author{Muqing Cao*, Xinhang Xu*, Shenghai Yuan, Kun Cao, Kangcheng Liu, and~Lihua~Xie,~\IEEEmembership{Fellow,~IEEE}
\thanks{*Equal Contribution}
\thanks{All authors are with the School of Electrical and Electronic Engineering,
        Nanyang Technological University, 50 Nanyang Avenue, Singapore
        }
\thanks{$^{1}$ Corresponding author}%
}

\begin{document}

\maketitle
\thispagestyle{empty}
\pagestyle{empty}

\begin{abstract}
In this paper, we present the dynamic model and control of DoubleBee, a novel hybrid aerial-ground vehicle consisting of two propellers mounted on tilting servo motors and two motor-driven wheels.
DoubleBee exploits the high energy efficiency of a bicopter configuration in aerial mode, and enjoys the low power consumption of a two-wheel self-balancing robot on the ground.
Furthermore, the propeller thrusts act as additional control inputs on the ground, enabling a novel decoupled control scheme where the attitude of the robot is controlled using thrusts and the translational motion is realized using wheels.
A prototype of DoubleBee is constructed using commercially available components. 
The power efficiency and the control performance of the robot are verified through comprehensive experiments.
Challenging tasks in indoor and outdoor environments demonstrate the capability of DoubleBee to traverse unstructured environments, fly over and move under barriers, and climb steep and rough terrains.

\end{abstract}

\section*{Supplementary Material}
$\textbf{Video}$: \href{https://youtu.be/hcw4GKmW_vs}{https://youtu.be/hcw4GKmW\_vs}

\section{Introduction}
Over the past years, unmanned systems like UGVs and UAVs have been increasingly applied for various fields\cite{yuan2021survey}, such as item delivery\cite{er2013development}, agricultural spraying\cite{herwitz2004imaging}, structural maintenance\cite{cao2021distributed} and exploration of unknown region\cite{turtle2007dragonfly}. Thanks to their teleoperability and partial autonomy, more and more unmanned systems are used in interstellar explorations\cite{turtle2007dragonfly,di2020geospatial,47584_2017}. However, due to the energy inefficiency of multi-rotor UAVs and the low traversability of UGVs in unstructured environments, the operating environment of unmanned systems has many limitations. In response to these problems, researchers have proposed hybrid aerial-ground robots, which combine the high flexibility of UAVs and the long endurance of UGVs to face more complex unmanned tasks. According to the ground-drive systems employed, the existing hybrid aerial-ground robots can be classified into four kinds: active wheel-based, passive wheel-based, leg based, and self-rolling based.

The active wheel-based hybrid robots \cite{active1,active2,active3,activewheel4} are equipped with separate sets of actuators for air and ground locomotion.
They can be considered as standard multi-rotor UAVs with additional driving mechanisms such as motors or variable speed gears to generate torques for driving the wheels. 
These robots have excellent energy efficiency on the ground as the torques are applied directly to drive the wheels in the direction of movements.
However, 
the efficiency of the robots in aerial mode is low due to the need to carry the redundant weight of the additional driving systems.

\begin{figure}[]
    \centering
    \includegraphics[trim={0cm 0 0 0}, clip, width=1.0\linewidth]{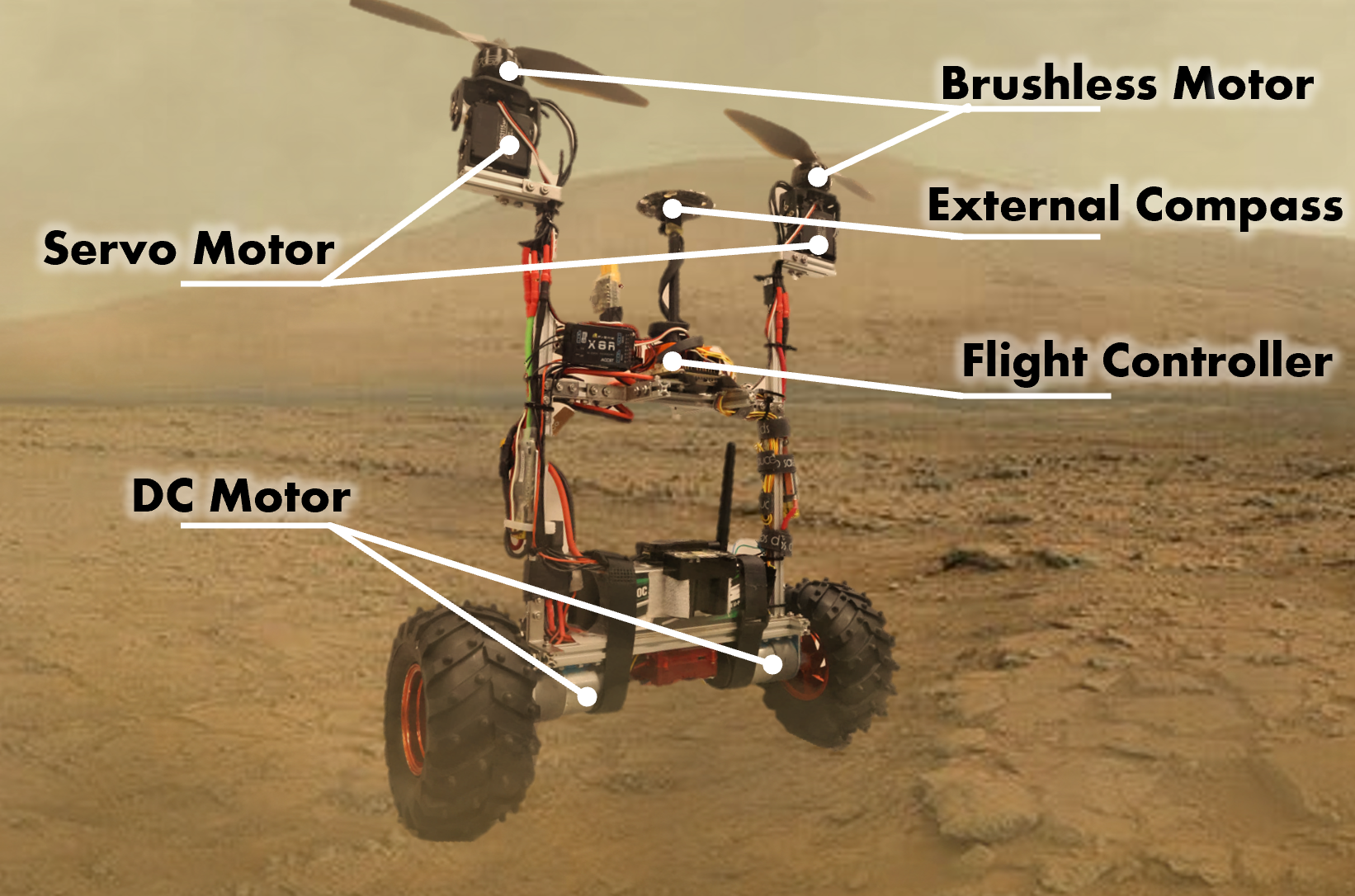}
    \caption{Our DoubleBee robot prototype.}
    \label{fig: doubleBee}    
\end{figure}

The passive wheel-based robots \cite{passivewheel1,passivewheel2,passivewheel3,passivewheel4,passivewheel5,passivewheel6} are equipped with free-rolling wheels.
On the ground, the robots adjust their attitude to produce thrusts in the intended direction of movement.
Compared to active wheel-based robots, the 
 lightweight passive wheels incur less energy loss in the air, and the mechanical structure is streamlined without significantly affecting the dynamics of the robot. 
 However, the energy efficiency during ground movement is comparatively lower because the direction of the thrusts generally does not align with the intended direction of movement. 
 Moreover, the translational movements on the ground are heavily coupled with the attitude changes of the robots, 
causing difficulty in generating stable video streams for exploration tasks.

Leg based robots based on bionic principles\cite{leg1,leg2} have been investigated, which use mechanical legs for ground locomotion. 
Compared with the first two types, complex tasks such as running and skateboarding can be completed using legs. 
Also, thanks to the flexibility of the legs, these robots can move easily on uneven terrains.
However, similar to active wheel-based robots, the complex mechanical structure of the legs brings a burden to the controller design and adds redundant weight in aerial mode. 
Moreover, the efficiency of the wheel-based robot on relatively level terrains is much higher than that of the leg-based robot. 

In addition, researchers have learned from the collision-resilient robot\cite{roll3} to design self-rolling based robots\cite{roll2,jia2022airframe}, which rotate their entire body as a wheel for ground locomotion.
Compared to the passive wheel-based robots, such an approach improves the ground efficiency but requires continuous attitude changes to move.
This poses challenges to the controller design and there is a risk of control failure in more complex environments.

In this work, we aim to design and build a hybrid aerial-ground robot that achieves a good balance between aerial and ground efficiency, and high traversability in rough terrains.
For aerial locomotion, we adopt a bicopter setup, which is the most energy-efficient type of uniaxial multi-rotor configuration \cite{qin2020gemini}.
To achieve good ground efficiency without adding significant air load, 
we adopt a minimal setup of two active wheels for ground locomotion.
The resulting robot (Figure \ref{fig: doubleBee}) can be considered a combination of a two-wheel self-balancing robot (balance bot) and a bicopter, hence the name DoubleBee.
Such a configuration brings various benefits.
First, 
thanks to the similarity between the dynamics and the control of bicopter and balance bot, a simple but effective control method enables a smooth transition between the aerial and the ground mode.
Second, the propeller thrusts provide additional control inputs on the ground, enabling a novel decoupled control strategy, where the pitch angle of the robot is controlled by the thrusts and the translational movements are controlled by the wheels.
The decoupled mode greatly extends the ground capability of the robot, allowing the robot to rise to an upright position from a static lying pose, passes under low barriers, and traverse rough terrains with a desired pitch.
All these tasks are not possible for a traditional balance bot, whose angular and translational movements are tightly coupled. 
Figure \ref{fig: MEI} shows a sequence of maneuvers executed by DoubleBee. 
The main contributions of the paper are summarized as:
\begin{figure*}[]
    \centering
    \includegraphics[trim={0 0cm 0cm 0cm}, clip, width=0.9\linewidth]{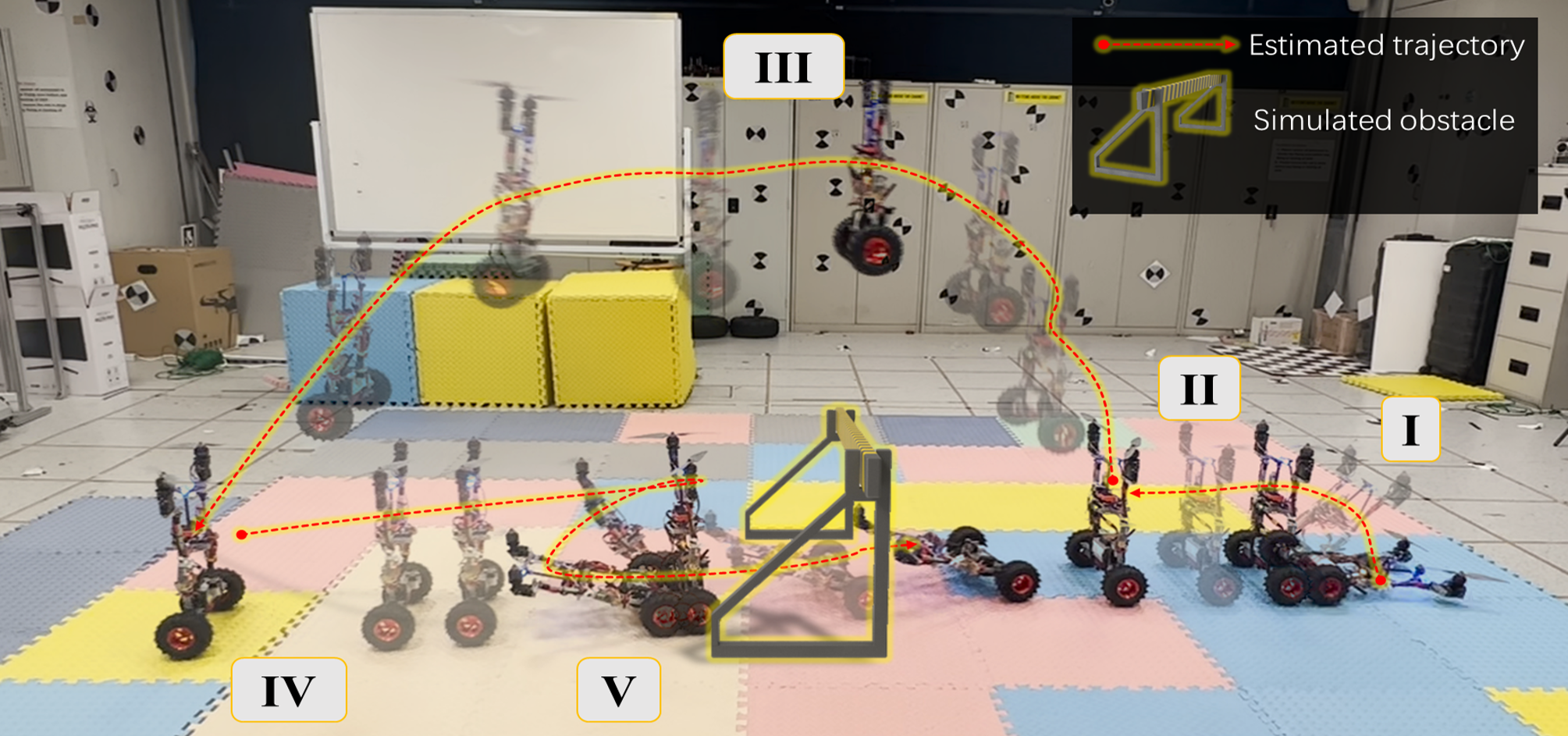}
    \caption{A demonstration of DoubleBee flying over and then passing under a simulated barrier in different control modes: (I) standing up in decoupled mode; (II) moving in ground mode; (III) flying in aerial mode; (IV) landing and moving in ground mode; (IV) leaning back in decoupled mode and moving forward.}
    \label{fig: MEI}    
\end{figure*}
\begin{itemize}
    \item A novel hybrid aerial-ground robot with two propellers and two active wheels is proposed and built; 
    \item Detailed dynamics modeling and controller design are carried out, including a smooth aerial-ground transition strategy and a novel decoupled control scheme;
    \item Experimental results verify the control performance in different control modes and the high energy efficiency of DoubleBee;
    \item Challenging indoor and outdoor missions are carried out to showcase the capability of DoubleBee in crossing over and under obstacles and traversing steep terrains. 
\end{itemize}

The rest of the paper is organized as follows: Section \ref{sec: modelling} details the dynamic model of the robot.
Controller design for different control modes is explained in Section \ref{sec: control}.
Section \ref{sec: exp} describes the constructed robot platform and presents the experimental verification in terms of power consumption, control performance and execution of indoor and outdoor missions.
Section \ref{sec: conclu} concludes the paper.

\section{Dynamics Modelling}\label{sec: modelling}

\begin{table}[]
  \centering
\caption{Notation for dynamics modelling}
\label{tab: notation}
\def\arraystretch{1.2}
\begin{tabular}{|l|l|}
\hline
Symbol                                                              & Meaning                                                                                                                                                                                                                    \\ \hline
$\vb{\pos}^\inertial$                                               & Position of the CG of  the chassis                                                                                                                                                                                         \\ \hline
$\vb{\pos}_w^\inertial$                                             & Position of the CG of the wheel                                                                                                                                                                                            \\ \hline
$\mass, \mass_{w}$                                                  & Masses of the chassis and the wheel, respectively                                                                                                                                                                          \\ \hline
$\roll,\pitch,\yaw$                                                 & Roll, pitch, yaw Euler angle of the chassis                                                                                                                                                                                \\ \hline
$\angwheel_1,\angwheel_2$                                           & \begin{tabular}[c]{@{}l@{}}Rotation angles of the left and the right wheels \\ about the $y$-axis of the wheels\end{tabular}                                                                                               \\ \hline
$\tilt_{l},\tilt_{2}$                                               & \begin{tabular}[c]{@{}l@{}}Tilt angle of the servo motors with respect to the \\ body $z$-axis of the chassis\end{tabular}                                                                                                 \\ \hline
$\gbf{\angvel}^\body$                                               & Angular velocity of the chassis                                                                                                                                                                                            \\ \hline
$\gbf{\angvel}^\bodywheelone_{w1},\gbf{\angvel}^\bodywheeltwo_{w2}$ & \begin{tabular}[c]{@{}l@{}}Angular velocity of the wheels. When a wheel only \\ rotates about its axis of rotation, $\gbf{\angvel}^\bodywheelone=[0, 0, \dot{\angwheel}]^\top$\end{tabular}                                \\ \hline
$\screw{\cdot}$                                                     & Screw-symmetric cross product matrix of a vector                                                                                                                                                                           \\ \hline
$\rotmatrix_x(\cdot)$                                               & \begin{tabular}[c]{@{}l@{}}Rotation matrix obtained by rotating around the \\ $x$ axes by an angle\end{tabular}                                                                                                            \\ \hline
$\rotmatrix$                                                        & \begin{tabular}[c]{@{}l@{}}Rotation matrix from the inertial frame to \\ the body frame, $\rotmatrix=\rotmatrix_z(\yaw) \rotmatrix_y(\pitch) \rotmatrix_x(\roll)$\end{tabular}                                             \\ \hline
$\rotmatrix_{w1}, \rotmatrix_{w2}$                                  & \begin{tabular}[c]{@{}l@{}}Rotation matrix from the inertial frame to \\ the wheel-fixed frame, $\rotmatrix_{w1}=\rotmatrix_z(\yaw) \rotmatrix_x(\roll)$\end{tabular}                                                      \\ \hline
$\tor_{w1}, \tor_{w2}$                                              & Torque by the DC motors on the wheels                                                                                                                                                                                      \\ \hline
$\torvec_{w1}, \torvec_{w2}$                                        & $\torvec_{w1}=[0, \tor_{w1},0], \torvec_{w2}=[0,\tor_{w2},0]$.                                                                                                                                                             \\ \hline
$\thrust_1,\thrust_2$                                               & Thrusts produced by the left and the right propellers                                                                                                                                                                      \\ \hline
$\vb{\for}^\body_{1},\vb{\for}^\body_{2}$                           & \begin{tabular}[c]{@{}l@{}}Forces acting on the chassis due to thrusts, \\ $\vb{\for}^\body_1=\rotmatrix_y(\tilt_{1})[0, 0, \thrust_1]^\top$, $\vb{\for}^\body_1=\rotmatrix_y(\tilt_2)[0, 0, \thrust_2]^\top$\end{tabular} \\ \hline
$\vb{\for}^\body_{w1},\vb{\for}^\body_{w2}$                         & \begin{tabular}[c]{@{}l@{}}Forces acting on the chassis by the left and \\ the right wheels.  On level ground, \\ $\forcevec^\body_{w1}=[\horiz_1,0, \normal_1]^\top$.\end{tabular}                                        \\ \hline
$\vb{\for}^\inertial_g,\vb{\for}^\inertial_{wg}$                    & \begin{tabular}[c]{@{}l@{}}Gravity force of the main body and each wheel. \\ Both wheels are assumed to have the same weight.\end{tabular}                             \\ \hline
$\vb{\for}^\bodywheelone_{f1}, \vb{\for}^\bodywheeltwo_{f2}$        & \begin{tabular}[c]{@{}l@{}}Friction force acting on the wheels by the ground.\\ On a level ground, $\forcevec_{f1}^\bodywheelone=[\friction_1, 0, 0]^\top$.\end{tabular}                                                   \\ \hline
$\vb{\for}^\bodywheelone_{n1}, \vb{\for}^\bodywheeltwo_{n2}$        & \begin{tabular}[c]{@{}l@{}}Normal force acting on the wheels by the ground.\\ On a level ground, $\forcevec^\bodywheelone_{n1}=[0, 0, \normal_1]^\top$.\end{tabular}                                                         \\ \hline
$\moi^\body,\moi^\bodywheel_{w}$                                    & Moment of inertial matrices of the chassis and wheels                                                                                                                                                                      \\ \hline
$\moicombined^\body$                                                & Moment of inertial of the entire robot                                                                                                                                                                                     \\ \hline
$\moiscaler_y, \moiscaler_{wy}$                                     & \begin{tabular}[c]{@{}l@{}}Moment of inertial of the chassis and the wheels \\ about their respective body $y$-axes\end{tabular}                                                                                           \\ \hline
$\dist_y, \dist_z$                                                  & \begin{tabular}[c]{@{}l@{}}Distance between the CG of the chassis and the \\ center of  the propeller along the body frame \\ $y$ and $z$ axis\end{tabular}                                                                \\ \hline
$\dist_{wy}, \dist_{wz}$                                            & \begin{tabular}[c]{@{}l@{}}Distance between the CG of the chassis and the \\ wheel along the body frame $y$ and $z$ axis\end{tabular}                                                                                      \\ \hline
$\radius$                                                           & Radius of the wheels                                                                                                                                                                                                       \\ \hline
$\distvec^\body_1, \distvec^\body_2$                                & \begin{tabular}[c]{@{}l@{}}Distance vector from the CG of the chassis to the \\ propellers, $\distvec^\body_1=[0, \dist_y, \dist_z]^\top$\end{tabular}                                                                     \\ \hline
$\distvec^\body_{w1}, \distvec^\body_{w2}$                          & \begin{tabular}[c]{@{}l@{}}Distance vector from the CG of the chassis to the\\  wheels, $\distvec^\body_{w1}=[0, \dist_{wy}, -\dist_{wz}]^\top$\end{tabular}                                                               \\ \hline
$\distvec^\bodywheelone_{c1}, \distvec^\bodywheeltwo_{c2}$          & \begin{tabular}[c]{@{}l@{}}Distance vector from the CG of wheels to the \\ contact point with the ground. \\ On a level ground, $\distvec^\bodywheelone_{c1}=[0,0,-r]$\end{tabular}                                        \\ \hline
$\friction_1,\friction_2$                                           & \begin{tabular}[c]{@{}l@{}}Friction on the wheel against the movement of the\\  wheel.\end{tabular}                                                                                                                        \\ \hline
$\normal_1, \normal_2$                                              & Magnitude of normal force acting on the wheel.                                                                                                                                                                             \\ \hline
$\horiz_1,\horiz_2$                                                 & \begin{tabular}[c]{@{}l@{}}Horizontal component of the force acting on the \\ chassis  by the wheels.\end{tabular}                                                                                                         \\ \hline
$x, x_{wi}$                                                         & \begin{tabular}[c]{@{}l@{}}Displacement of the CG of the chassis and \\ the wheels in the $x$-axis\end{tabular}                                                                                                            \\ \hline
\end{tabular}
\end{table}

\begin{figure*}[htbp]
        \centering
        \includegraphics[width=0.9\linewidth]{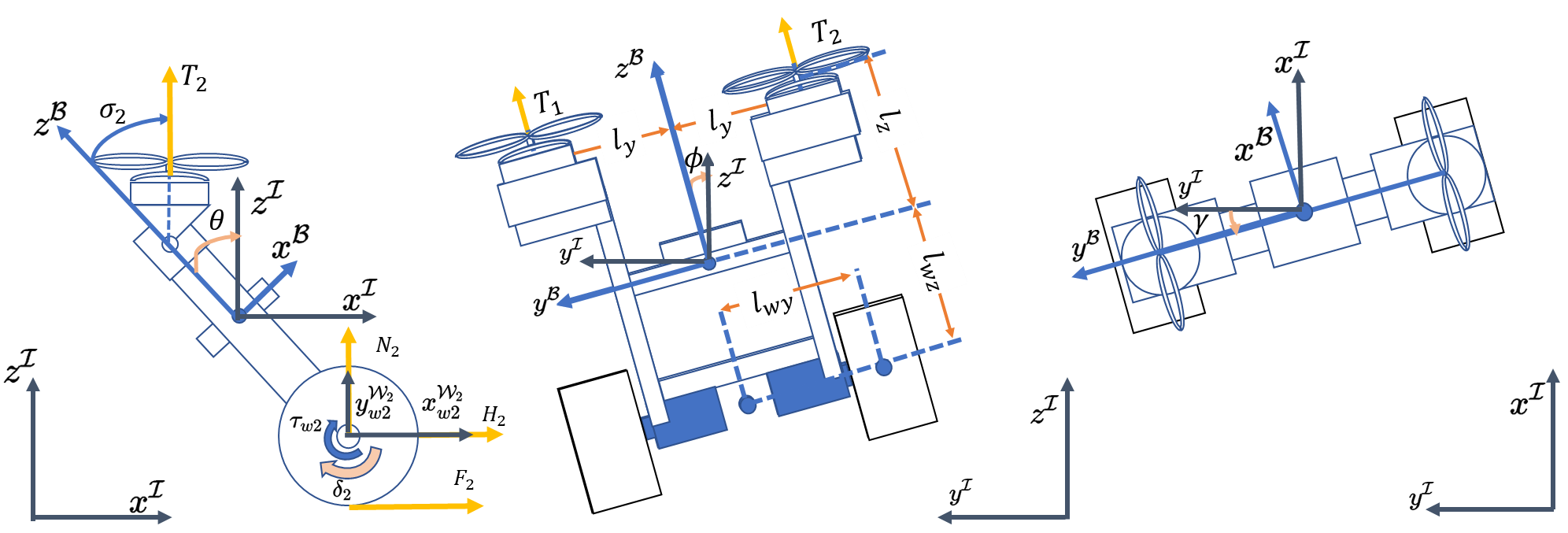}
        \caption{The body frame of the chassis $\{x^\body,y^\body,z^\body\}$, the right wheel frames $\{x^\bodywheeltwo_{w2},y^\bodywheeltwo_{w2},z^\bodywheeltwo_{w2}\}$ and the inertial frame $\{x^\inertial,y^\inertial,z^\inertial\}$.}
        \label{fig: frames}        
    \end{figure*}

For ease of presentation, the frequently used symbols are presented in Table \ref{tab: notation}. 
To present the full dynamics of the system, we independently analyze the dynamics of the chassis of the robot (everything of the robot except the two wheels) and the wheels.
Throughout the paper, we use the superscripts $\inertial$, $\body$, $\bodywheelone$, and $\bodywheeltwo$ to indicate that a vector is in the inertial frame, the body frame  of the chassis and the fixed frames of the left and the right wheels\footnote{the fixed frame of a wheel has the origin at the center of the wheel, $x$ axis always pointing at the direction of the forward movement and $y$ axis is the rotation axis.}, respectively.
These frames are shown in Figure \ref{fig: frames}.
The dynamic model is detailed in the following equations:
\begin{align}
&\mass\ddot{\posvec}^\inertial=\vb{\for}^\inertial_g+\rotmatrix(\vb{\for}^\body_1+\vb{\for}^\body_2+\vb{\for}^\body_{w1}+\vb{\for}^\body_{w2}),\label{eq: forcebody}\\
&\moi^\body\dot{\angvelvec}^\body+\screw{\angvelvec^\body}\moi^\body\angvelvec^\body = -(\gbf{\tor}_{w1}+\gbf{\tor}_{w2})+\screw{\distvec_{w1}^\body}\gbf{\for}^\body_{w1}+\nonumber\\
&\quad\quad\quad\quad\screw{\distvec_{w2}^\body}\gbf{\for}^\body_{w2}+
\screw{\distvec_1^\body}\vb{\for}^\body_1+\screw{\distvec_2^\body}\vb{\for}^\body_2,\label{eq: torquebody}\\
&\mass_{w}\ddot{\vb{\pos}}^\inertial_{wi} = \vb{\for}^\inertial_{wg}+\rotmatrix_w(\vb{\for}^{\bodywheel_i}_{fi}+\vb{\for}^{\bodywheel_i}_{ni})-\rotmatrix\vb{\for}^\body_{wi},\label{eq: forcewheel}\\
&\moi_w^{\bodywheel}\dot{\gbf{\angvel}}^{\bodywheel_i}_{w}+\screw{\angvelvec^{\bodywheel_i}_{wi}}\moi\angvelvec^{\bodywheel_i}_{wi}=\torvec_{wi}+\screw{\distvec^{\bodywheel_i}_{ci}}\forcevec^{\bodywheel_i}_{fi}\label{eq: torquewheel},\\
&\ddot{\posvec}^\inertial_{wi}=\ddot{\posvec}^\inertial+\screw{\dot{\angvelvec}^\body}\distvec^\body_{wi},\label{eq: acc}
\end{align}
$\forall i\in\{1,2\}$.
Equations \eqref{eq: forcebody} and \eqref{eq: torquebody} detail the translational and rotational dynamics of the main chassis, while equations \eqref{eq: forcewheel} and \eqref{eq: torquewheel} describe the translational and rotational dynamics of the wheels.
Equation \eqref{eq: acc} shows the kinematic relationship between the acceleration of the wheels and that of the chassis.
Some assumptions are made in the above equations:
(i) the normal force $\forcevec^{\bodywheel_i}_{ni}$ and the reaction force of $\forcevec^\body_{wi}$ pass through the center of gravity (CG) of the wheel, so they do not generate any torques; 
(ii) the torques generated by the servo motors are ignored; 
(iii) the change in the distance vectors from the CG of the chassis to the propellers induced by the tilting servo motors is ignored.

\subsection{Aerial Mode}
When the robot is airborne, both the friction forces and the contact forces on the wheels, $\forcevec_{n1}^\bodywheelone$, $\forcevec_{n2}^\bodywheeltwo$, $\forcevec_{f1}^\bodywheelone$ and $\forcevec_{f2}^\bodywheeltwo$ are zero. 
Suppose that the DC motors apply torques to the wheels such that both wheels remain static relative to the chassis.
The entire robot becomes a rigid body and the dynamics equations are simplified to those of a standard bicopter \cite{qin2020gemini}:
\begin{align}
    &(\mass+\mass_w)\ddot{\posvec}^\inertial=\forcevec^\inertial_g+2\forcevec^\inertial_{wg}+\rotmatrix(\forcevec_1^\body+\forcevec_2^\body),\label{eq: coptforce}\\
    &\moicombined^\body\dot\angvelvec^\body+\screw{\angvelvec^\body}\moicombined^\body\angvelvec^\body=\screw{\distvec_1^\body}\vb{\for}^\body_1+\screw{\distvec_2^\body}\vb{\for}^\body_2.\label{eq: copttorq}
\end{align}
Some simplifications are made in the above equations.
The body frame of the entire robot is assumed to coincide with that of the chassis, and the center of gravity of the entire robot is assumed to be close to that of the chassis.
Otherwise, the angular velocity and the distance vectors in Equation \eqref{eq: copttorq} should be updated to the new reference frame and the new center of gravity.
In general, the mass of a wheel is small compared to the main chassis, which includes all the motors and the battery.
Therefore, the above approximation is reasonable.
\subsection{With Ground Contact}\label{subsec: ground model}
Consider that the robot is in contact with a level ground.
The ground provides horizontal friction and vertical support forces to the robot, and the dynamics of the robot become similar to that of a two-wheel balance robot \cite{wu2012two}.
Of particular interest are the translational dynamics along the $x$-axis and the rotational dynamics about the $y$-axis (left diagram of Figure \ref{fig: frames}):

\begin{align}
    & \mass\ddot{x}=\horiz_1+\horiz_2+\thrust_1\sin(\pitch+\tilt_1)+\thrust_2\sin(\pitch+\tilt_2),\label{eq: transchassisx}\\
    &\moiscaler_y\ddot{\pitch}=\thrust_1\dist_z\sin{\tilt_1} + \thrust_2\dist_z\sin{\tilt_2} -(\horiz_1+\horiz_2)\dist_{wz}\cos{\pitch} \nonumber\\
    &\quad\quad\quad+(\normal_1+\normal_2)\dist_{wz}\sin{\pitch}- (\tor_{w1}+\tor_{w2}),\label{eq: rotchassisy}\\
    &\mass_w\ddot{x}_{wi}=\friction_i-\horiz_i,\label{eq:transwheelx}\\
    &\moiscaler_{wy}\ddot{\angwheel}_i=\tor_{wi}-\friction_i\radius,\label{eq: rotwheely}\\
    &\angwheel_i\radius = x_{wi},\label{eq: angletodisp}
\end{align}
$\forall i\in\{1,2\}$. Equations \eqref{eq: transchassisx} and \eqref{eq: rotchassisy} govern the translational and rotational dynamics of the chassis,
Equations \eqref{eq:transwheelx} and \eqref{eq: rotwheely} govern the translational and rotational dynamics of the wheels. Equation \eqref{eq: angletodisp} describes the kinematic relationship between the rotation of the wheels and the displacement in the $x$-axis, under the non-slip condition.
For a traditional balance bot, $\thrust_1=\thrust_2=0$. From the above equations, the forward movement of a balance bot requires the horizontal forces $\horiz_1$ and $\horiz_2$ provided by the wheels.
However, the horizontal forces also affect the pitch of the robot (Equation \eqref{eq: rotchassisy}), 
resulting in a coupling between the position and attitude of the robot.
The addition of the propeller thrusts $\thrust_1$ and $\thrust_2$ contributes to both the translational and angular movements of the robot.
From Equation \eqref{eq: rotchassisy}, when $\pitch<0$ and $\tilt_1=\tilt_2>0$, a positive pitch angular acceleration can be produced by applying a dominant torque through propeller thrust. 
To achieve a negative angular acceleration, note that when the robot is static or moving slowly and the thrusts are low, a dominant negative torque is produced by the normal support force $\normal_1$ and $\normal_2$.
Similarly, when $\pitch<0$ and $\tilt_1=\tilt_2>0$, positive and negative pitch acceleration can be produced by reducing and increasing the thrust, respectively.
In Equation \eqref{eq: transchassisx}, 
when the tilt angles of the motors are close to the negative pitch angle, i.e., $\tilt_1=\tilt_2\approx-\pitch$, the effect of the thrusts is negligible and the horizontal movement of the robot mainly depends on the horizontal forces $\horiz_1$ and $\horiz_2$.
Thus, such a model allows for decoupling between the position and attitude of the robot.

\section{Control Design}\label{sec: control}
Four types of control modes are introduced in this section: the aerial mode, the ground mode, the aerial-ground transition, and the decoupled mode.
\subsection{Aerial Mode}\label{subsec: aerial control}
In strong resemblance to the standard bicopter\cite{qin2020gemini}, the DoubleBee's attitude controller in aerial mode uses a cascade PID design whose outer loop is a proportional controller to track the desired attitude $\rotmatrix_d$, which comes from the pilot or the position controller. The desired angular velocity is obtained as \\
\begin{align}
\angvelvecdesir^\body & =\airka\frac{1}{2}(\rotmatrix^\top_d\rotmatrix-\rotmatrix^\top\rotmatrix_d)^\vee,
\end{align}\\
where $^\vee$ is the mapping from $so(3)$ to $\mathbb{R}^3$, $\airka$ is the diagonal matrix with the diagonal terms as proportional gains for roll, pitch, and yaw control. The inner-loop PID controller tracks the desired angular velocity: \\
\begin{equation}
\begin{aligned}
\angvelvec^\body_e & =\angvelvecdesir^\body-\angvelvec^\body,\\
\velcontrol & =\airkp \angvelvec^\body_e +\airki \int \angvelvec^\body_e+\airkd  \dot{\angvelvec}^\body_e,
\end{aligned}
\end{equation}
where $\angvelvec_e$, $\velcontrol$ are the angular velocity error and the control signal output by the PID controller, respectively. $\airkp,\airki,\airkd$ represent the diagonal matrices consisting of the proportional, integral, and differential gains, respectively.
A mixer computes the output of motors and servo motors as:
\begin{equation}
\left[\begin{array}{c}
T_1 \\
T_2 \\
\sigma_1 \\
\sigma_2
\end{array}\right]=\left[\begin{array}{cccc}
1 & 0 & 0 & 1 \\
-1 & 0 & 0 & 1 \\
0 & 1 & -1 & 0 \\
0 & 1 & 1 & 0
\end{array}\right]\left[\begin{array}{c}
\velcontrol \\
\thrin
\end{array}\right],
\label{bicopter}
\end{equation}
where $\thrin$ is the desired level of throttle from the pilot input or the position controller. The mixer matrix shows that the roll angle is controlled by the differential thrust of $T_1$ and $T_2$, and the control of the pitch and yaw are handled by the tilting of both motors.

\subsection{Ground Mode}\label{subsec: ground control}
DoubleBee's control scheme is similar to that of a balance bot when it is running in ground mode. Using a cascade controller, the outer loop tracks the desired velocity along the body $x$-axis and the inner loop tracks the desired pitch angle of the robot \cite{Philip2020balance}. The workflow can be written as:
\begin{equation}
\begin{aligned}
\pitch_e & =\balancekv(\wheelveldesired-\wheelvel)-\pitch,\\
\tau_{w1} & =\balancekp
\mathcal{\pitch}_e+\balanceki \int \pitch_e+\balancekd  \dot{\pitch}_e-\balanceks (\desiredsteer-\steerrate),\\
\tau_{w2} & =\balancekp
\mathcal{\pitch}_e+\balanceki \int \mathcal{\pitch}_e+\balancekd  \dot{\pitch}_e+\balanceks (\desiredsteer-\steerrate),
\end{aligned}
\label{balancebot}
\end{equation}\\
where $\wheelveldesired,\wheelvel,\desiredsteer,\steerrate$ are the expected and actual values of the vehicle velocity and the steer(yaw) rate, respectively. $\balancekv$ denotes the gain of the velocity control loop and the inner loop PID terms are defined as $\balancekp,\balanceki,\balancekd$ and $\balanceks$.  
\subsection{Aerial-Ground Transition}
Sections \ref{subsec: aerial control} and \ref{subsec: ground control} show that the control inputs for correcting the pitch and yaw errors are the servo tilting angles in the aerial mode and the motor torques in the ground mode.
Activating both control inputs simultaneously may result in strong oscillations in the attitude.
On the other hand, a direct switch from one mode to another may cause the control input from the previous mode to become an initial disturbance.
To realize a smooth transition from ground to air and vice versa, an input scaling is implemented that gradually reduces the control input in one mode and increases the control input in the other. 
The scaling is given as:
\begin{equation}
\begin{aligned}
\sigma_1 & =\scale \sigma_1^a, \\
\sigma_2 & =\scale \sigma_2^a ,\\
\tor_{w1} & =(1-\scale) \tau_{w1}^b ,\\
\tor_{w2} & =(1-\scale) \tau_{w2}^b ,\\
\end{aligned}
\end{equation}
where $\sigma_1^a,\sigma_2^a$, $\tau_{w1}^b$, and $\tau_{w2}^b$ represent the original control commands computed from the controllers in aerial mode and ground mode. The scaling factor $\scale$ is computed in a piece-wise linear fashion as:
\begin{equation}
\scale =
\begin{cases}
0,&\ \thrin \leq \thridle\\ 
\frac{(\thrin-\thridle)}{\thrhover-\thrin},&\ \thridle<\thrin \leq \thrhover \\ 
1, & \text{otherwise}
\end{cases}
\end{equation}
where the $\thridle, \thrhover$ are the level of throttle in the idle state and that needed to hover in the air, respectively.
\subsection{Decoupled Mode}
The decoupled mode can be switched from and to the ground mode under pilot command.
As explained in Section \ref{subsec: ground model}, when the robot has ground contact, a decoupled control scheme can be realized, where the pitch control is handled by the propeller thrusts and tilting angles, and the position control is handled by the motor torques.
First, the desired angular velocity and the angular velocity error are computed:
\begin{equation}
\begin{aligned}
\desiredpitchrate&=\decouplek_p(\pitch_d-\pitch),\\
\uncouplethetae & = \desiredpitchrate-\dot{\pitch},
\end{aligned}
\end{equation}
where $\decouplek_p$ is the gain of the proportional controller.
As explained in Section \ref{subsec: ground model},
depending on the sign of the pitch angle, the thrusts may be increased or decreased to generate the desired torque. 
Hence, the following rule is introduced to compute the error in throttle level:
\begin{equation}
\thrusterror=
\begin{cases}
-\uncouplethetae,&\ \pitch>0 \text{ and } \dot{\pitch} \leq \dot{\pitch}_d <0 \text{ or}\\ 
&\ \pitch<0 \text{ and not } \dot{\pitch} \geq \dot{\pitch}_d>0   \\ 
\uncouplethetae.& \text{otherwise}
\end{cases}
\end{equation}
The throttle outputs are the sum of a reference throttle level and the output of a PID controller:
\begin{equation}
\begin{aligned}
\thrust_1=\thrust_2=\thrust^{hold}+\decouplekTp \thrusterror + \decouplekTi \int \thrusterror +\decouplekTd  \dot{\thrusterror}, 
\end{aligned}
\end{equation}
where $\thrust^{hold}$ is the level of throttle capable of lifting the robot from a completely flat status ($\pitch\geq\pm\frac{\pi}{2}$).
From our analysis in Section \ref{subsec: ground model}, the tilting angles should be set close to the negative pitch $-\pitch$ to minimize the effect on the translational movement of the robot.
In practice, the tilting angles are computed as the sum of the reference value and a bias obtained from a PID controller:
\begin{equation}
\begin{aligned}
\tilt_1=\tilt_2=-\pitch+\decoupleksigmap \uncouplethetae+ \decoupleksigmai \int \uncouplethetae +\decoupleksigmad  {\uncouplethetaedot}.
\end{aligned}
\end{equation}
The bias term increases the torque produced by the propeller thrusts. This is especially necessary when the robot is in the upright status ($\pitch\approx0$) because the lever arm for the thrust to produce torque is very small (See Equation \eqref{eq: rotchassisy}).
Finally, the motor torques are computed from the desired horizontal velocity and the desired steer rate:
\begin{equation}
\begin{aligned}
\tau_{w1} &= \decouplekv \wheelveldesired - \decoupleks \desiredsteer,\\
\tau_{w2} &= \decouplekv \wheelveldesired + \decoupleks \desiredsteer,
\end{aligned}
\end{equation}
where the $\decouplekv,\decoupleks$ are the control gains.
\section{Experimental Verification}\label{sec: exp}
\subsection{Platform Construction}
A DoubleBee robot prototype is constructed as shown in Figure \ref{fig: doubleBee}. Two iFlight X$2814$ $880$KV brushless motors and APC $10\times4.5$ inch propellers are used for providing thrusts.
The brushless motors are mounted on two servo motors with maximum $15$Kg$\cdot$cm torque.
The ground drive system consists of two 30:1 metal gear motors and two $120\times60$mm wheels.
A Cube Orange flight controller provides the state estimation of the robot and generates control commands for the motors and servo motors.
The proposed controllers are integrated into the Ardupilot autopilot firmware as a new vehicle type.
The overall weight of the robot with a $5300$mAh $6$-cell lithium polymer battery is $2.78$Kg.
The size of the robot (excluding the propeller) is $51\times36\times12$cm.
\subsection{Power Consumption}
\begin{figure}[]
    \centering
    \includegraphics[trim={0.4cm 0 0 0}, clip, width=1.0\linewidth]{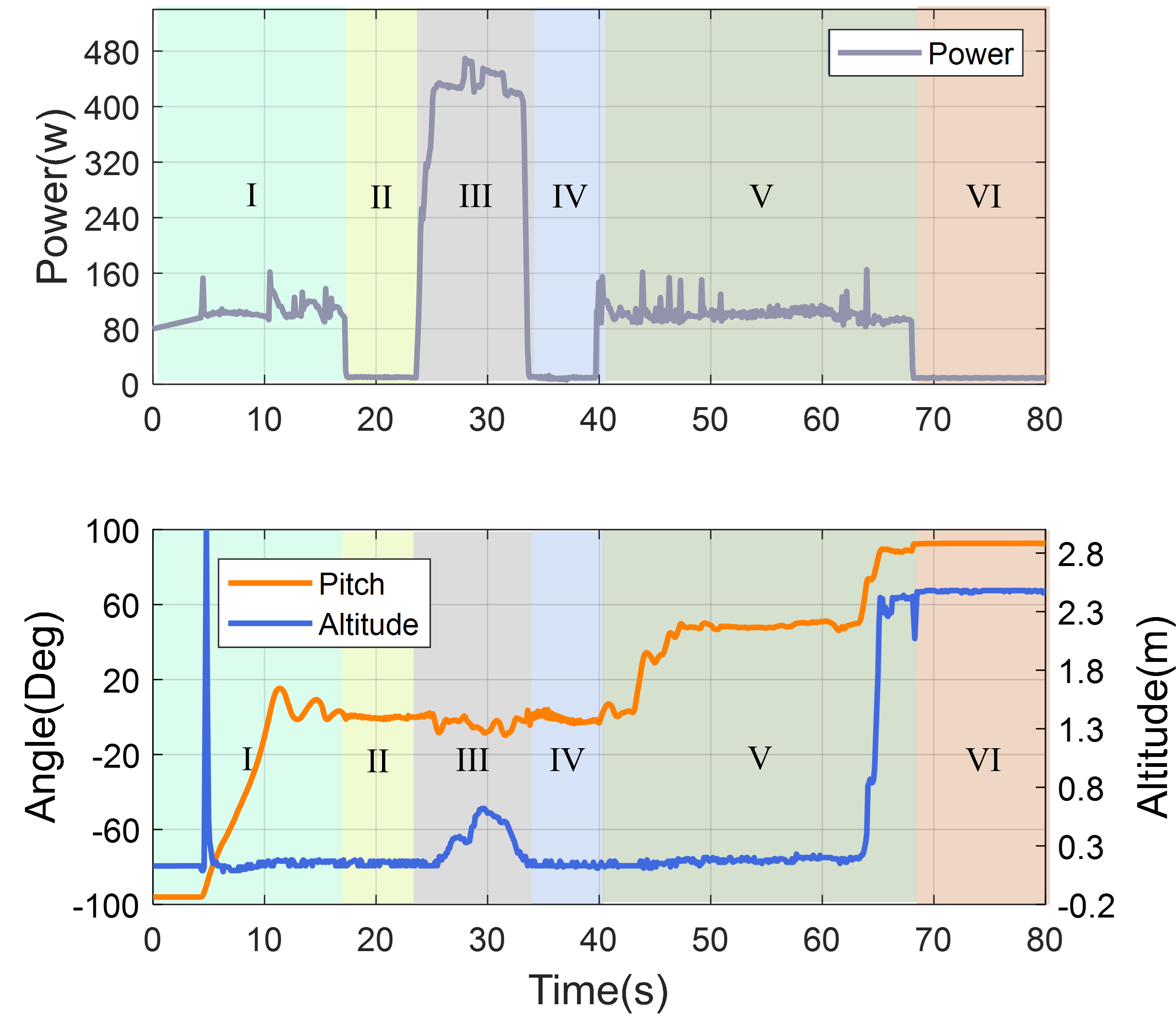}
    \caption{Energy consumption, altitude and pitch angle of the robot in a task consisting of different control modes.}
    \label{fig: energy}    
\end{figure}

\begin{figure}[]
    \centering
    \includegraphics[trim={0cm 0 0 0}, clip, width=1.0\linewidth]{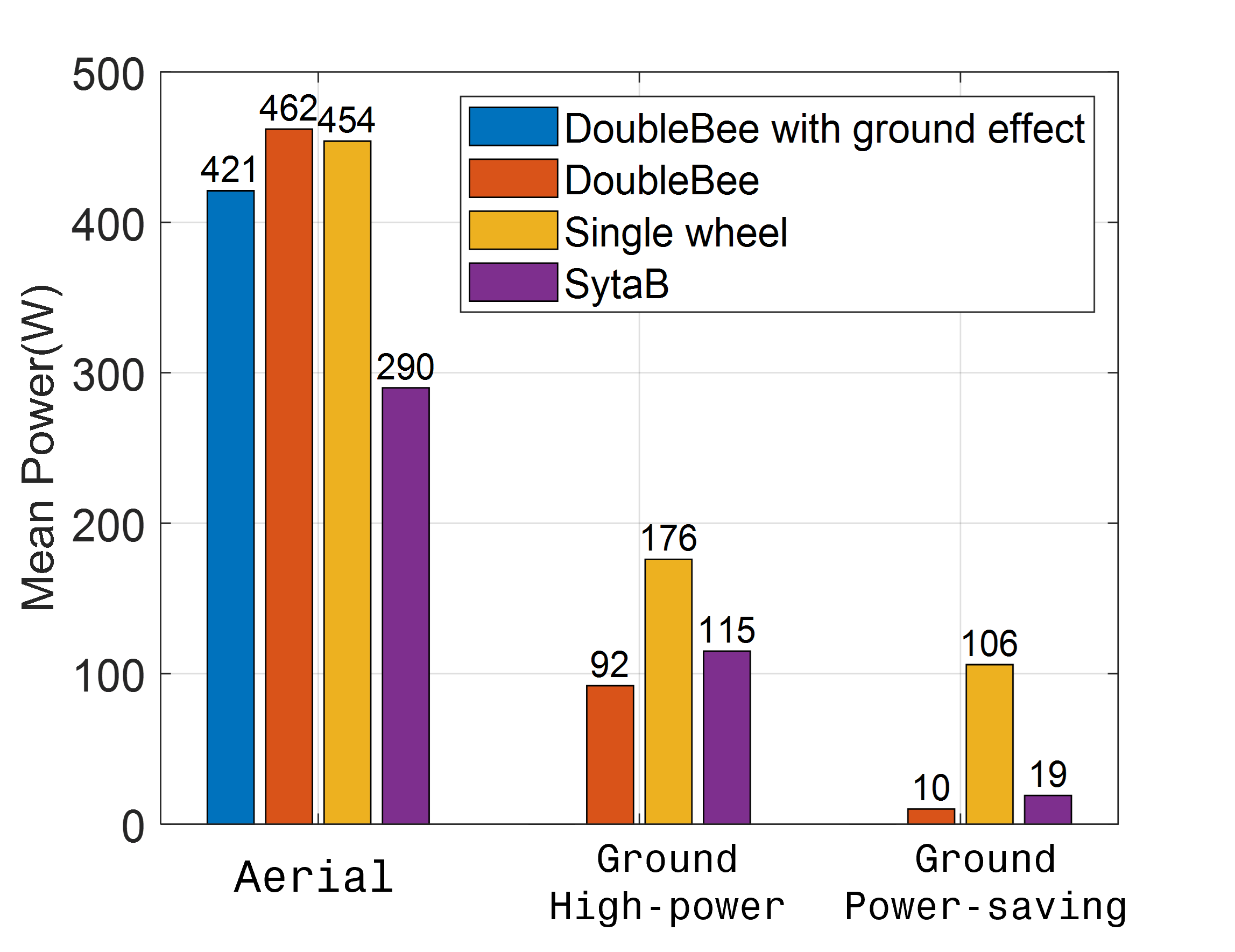}
    \caption{Comparison of average energy consumption in different modes. The average power consumption of the single-wheel robot is obtained directly from \cite{passivewheel6}, while the power consumption of SytaB is estimated from a figure in \cite{passivewheel2}.}
    \label{fig: energybar}    
\end{figure}

A test is conducted where the robot undergoes a sequence of control modes (Figure \ref{fig: MEI} and Figure \ref{fig: energy}): first, from a flat status ($\pitch\approx-90\degree$), the robot comes into the upright posture ($\pitch\approx0\degree$) in decoupled mode (I). Then, the robot switches to ground mode (II) for a short period, followed by taking off in aerial mode (III) (altitude changes can be observed in the bottom plot).
After a short flight, the robot lands and smoothly transits to ground mode again (IV), followed by switching to decoupled mode (V) and moving on the ground with an inclined angle ($\pitch\approx50\degree$).
Finally, the robot lies down completely in decoupled mode ($\pitch\approx90\degree$) and enters the idle state (VI).
High altitude estimates are observed in the first and the last phases because the laser beam emitted from the range sensor does not detect the ground when the robot is lying flat.
The power consumption during various phases can be observed in the top plot of Figure \ref{fig: energy}.
We compare the average power consumption of DoubleBee with two aerial-ground robots using the bi-copter configuration: SytaB \cite{passivewheel2} and a single passive wheel-based robot \cite{passivewheel6}.
Figure \ref{fig: energybar} shows the power consumption in different modes, where ground high-power mode indicates that more energy is consumed to ensure good traversability of the robots, 
and ground power-saving mode indicates the robot stays static or moves at a relatively slow speed.
For DoubleBee, we consider the decoupled mode to be the high-power mode and the ground mode to be  the power-saving mode.
In the air, DoubleBee consumes the most power when flying at an altitude where the ground effect is negligible (above 0.5m). 
Given that the weight of DoubleBee is $42\%$ higher than that of the single-wheel robot, the power-to-weight ratio of DoubleBee is lower, likely because the slim structure of DoubleBee reduces the drag induced by obstructing the downward airflow.
On the ground, DoubleBee has the lowest energy consumption in both high-power and power-saving modes.

\subsection{Ground Mode}
\begin{figure}[]
    \centering
    \includegraphics[width=1.05\linewidth]{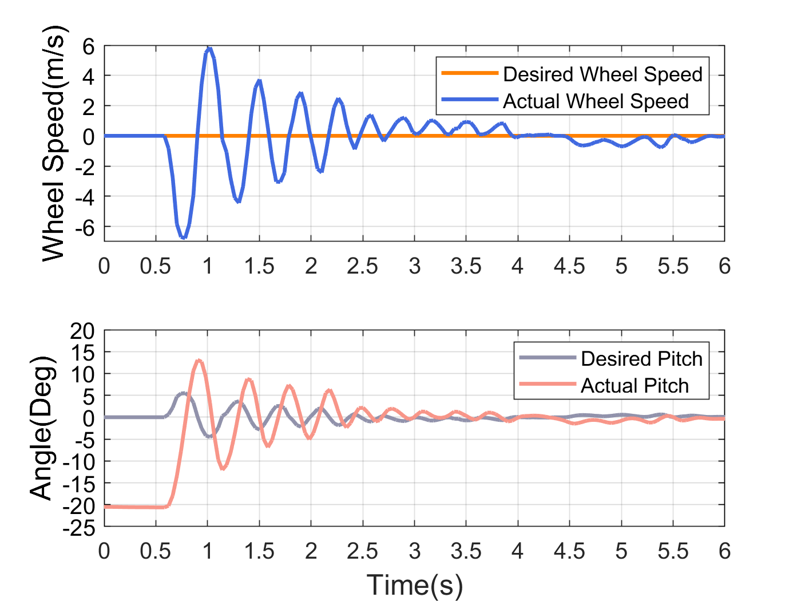}
    \caption{Wheel speed and pitch of the robot under a disturbance.}
    \label{fig: balancebot}    
\end{figure}
The control performance of the robot in ground mode is evaluated.
We simulate the response of the robot under a large disturbance by holding the robot at 
an initial pitch angle of around $-20\degree$ and then starting the controller. 
As seen in Figure \ref{fig: balancebot}, the pitch response overshoots initially but quickly converges the proximity of the $0\degree$.
In the long run, the pitch angle and the wheel speed oscillates in the proximity of zero, showing the stability of the system.
The small oscillations also show that continuous control commands are necessary to ensure that the robot maintains an upright configuration, which is a characteristic of a balance bot.

\begin{figure}[]
    \centering
    \includegraphics[width=1.05\linewidth]{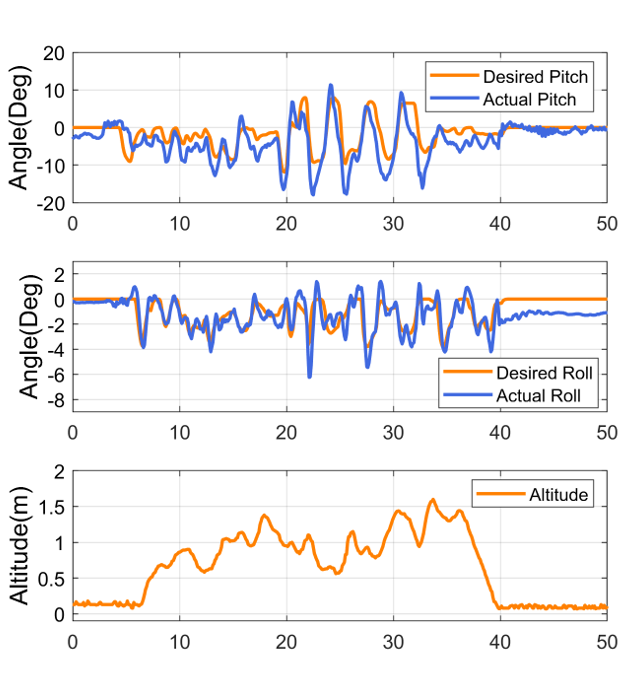}
    \caption{Attitude, desired attitude, and altitude of DoubleBee during the aerial-ground transition and manual flight.}
    \label{fig: airground}    
\end{figure}
\subsection{Aerial Mode and Aerial-ground Transition}
Figure \ref{fig: airground} illustrates the performance of the robot in aerial mode and aerial-ground transition. 
The robot initially stands upright in ground mode and the pilot increases the throttle and lifts the robot into the air.
During the flight, the pilot inputs attitude commands using the transmitter stick.
The bottom plot shows the altitude of the robot during the operation.
In the first two plots of Figure \ref{fig: airground}, it can be observed that both the roll and pitch follow the desired values well.
The pitch response is unable to reach large negative desired values, likely due to the CG of the robot being off-center.
Adjusting the battery placement in the body frame $x$-axis or providing a fixed bias (trim) to the servo tilt angle could alleviate the issue.
During the takeoff and landing stage, the attitude of the robot remains stable without large oscillations, showing that the proposed transition strategy enables the robot to transit smoothly from air to ground and vice versa.
\subsection{Decoupled Mode}
\begin{figure}[]
    \centering
    \includegraphics[width=1.0\linewidth]{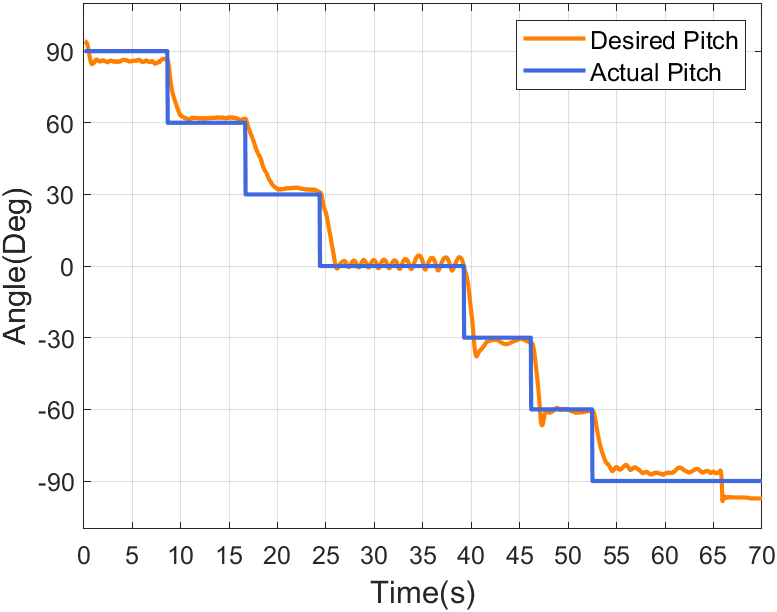}
    \caption{Pitch response in the decoupled mode under step commands.}
    \label{fig: decouple_stastic}    
\end{figure}

Two tests are conducted to evaluate the control performance in the decoupled mode. 
In the first experiment, the robot starts from an idle state and receives a sequence of pitch commands, each separated by $30\degree$.
Figure \ref{fig: decouple_stastic} shows that the actual pitch is able to reach the desired values and maintain stability at every desired value.
The pitch responses when the robot is rising from $\pitch=90\degree$ to $\theta=0\degree$ and descending from $\theta=0\degree$ to $\pitch=-90\degree$ are different: when rising, the pitch reaches the desired value more slowly without overshoot; when descending, the pitch responds quickly with overshoot.
This is because gravity is constantly pulling the robot downwards; it acts as extra damping when the robot is rising.
Small oscillations ($\pm3\degree$) are observed around $0\degree$. This is because the effect of gravity on rotation is negligible in the upright state (Equation \eqref{eq: rotchassisy}), thus the gain values tuned for good performance in inclined states are too large for this state.
A gain-scheduled controller can be implemented to achieve better overall performance.

\begin{figure}[]
    \centering
    \includegraphics[width=1.0\linewidth]{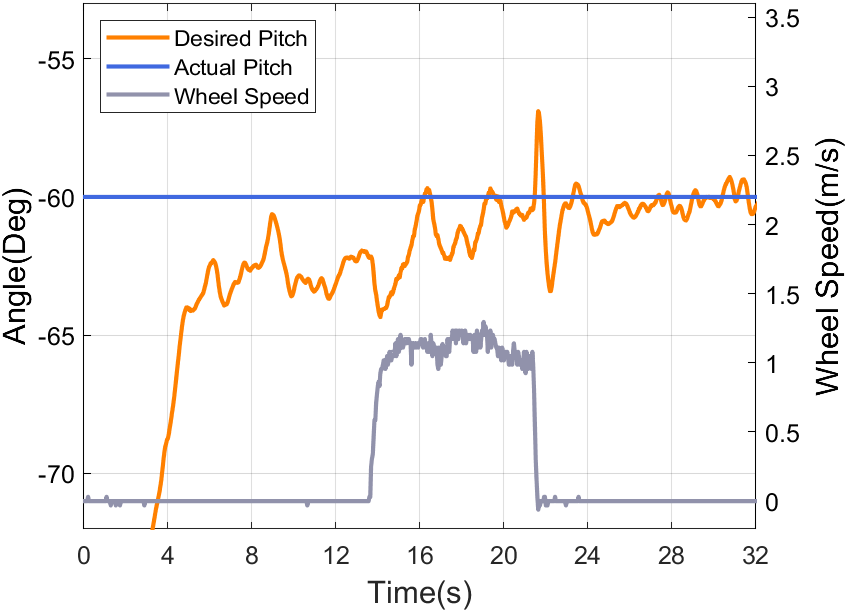}
    \caption{Pitch response in the decoupled mode during a forward translational movement.}
    \label{fig: decouple_run}    
\end{figure}

The second test evaluates the robustness of the pitch control under non-zero motor torques.
As observed from Figure \ref{fig: decouple_run},
the robot holds a desired pitch at $-60\degree$ when a forward movement is commanded.
When the wheels start rolling, a drop of $2.5\degree$ is observed, but the pitch angle quickly increases and reaches the desired value in $2.5$ seconds.
During the movement, the pitch is able to maintain within $3\degree$ of the desired value.
At the moment the rolling stops, an oscillation of $\pm3\degree$ is observed, but it settles in less than $2$ seconds.
Hence, the proposed decoupled control scheme enables robust control of the pitch angle under mild motor inputs.
Thus, the robot is capable of conducting a mission at an inclined angle.

\subsection{Challenging Tasks}
\begin{figure}[]
    \centering
    \includegraphics[width=1.0\linewidth]{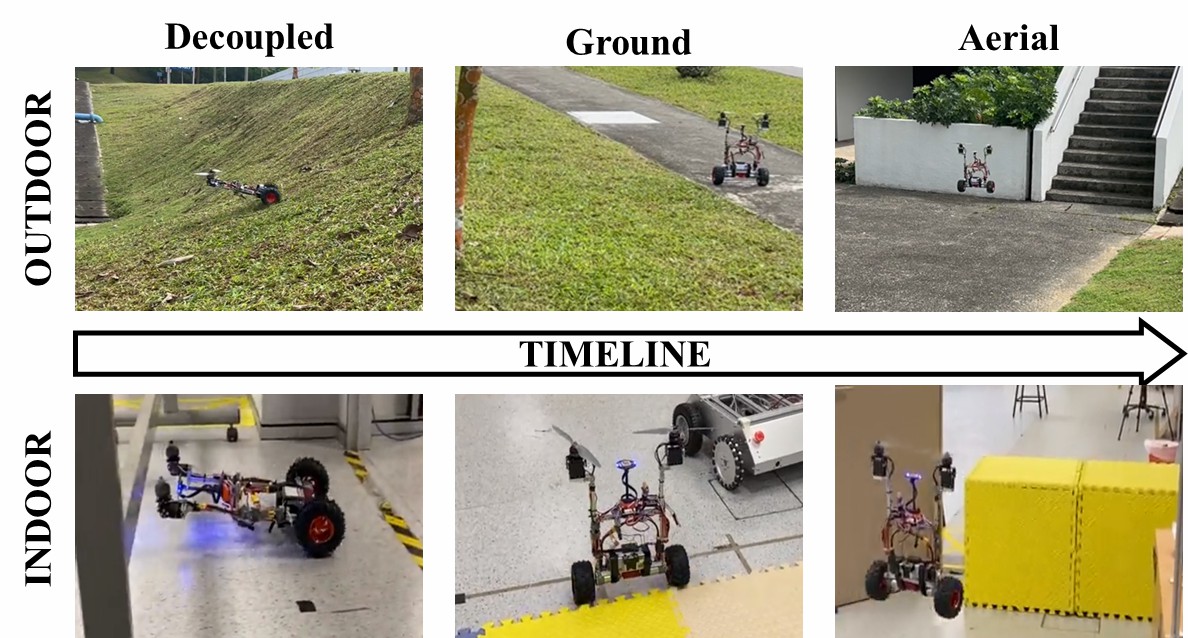}
    \caption{Illustrations of the challenging indoor and outdoor tasks.}
    \label{fig: exp}    
\end{figure}
We conduct two challenging tasks in which the robot is controlled to traverse regions with different types of obstacles.
In an indoor mission, the robot crosses under a low barrier, maneuvers around a fence, flies over an obstacle and through a door.
In an outdoor mission, the robot climbs up a steep grass field in decoupled mode before taking off on a slope and flying across a small cliff.
Figure \ref{fig: exp} illustrates these missions, 
and a video recording can be viewed in the supplementary material.
Such complex missions clearly demonstrate DoubleBee's capability to traverse rough terrain with varying elevations and effectively overcome obstacles encountered along the way.
\section{Conclusion}\label{sec: conclu}
In this work, we have implemented a novel and efficient aerial-ground robot named DoubleBee, which consists of two active wheels and two motor-driven propellers mounted on tilting servo motors. We have elaborated on its dynamics model and control methods and designed a series of indoor and outdoor experiments to demonstrate its multimodal motion capability and energy efficiency. 
Unlike most previous works, the DoubleBee improves the under-actuated characteristics of the balance bot, enabling it to accomplish decoupled control of speed and attitude and obtain better performance on rough terrains.
Although satisfactory results have been obtained with PID control schemes, more complex tasks such as interplanetary exploration often require higher autonomy and system stability. Therefore, for the next step, a more robust control strategy, an autonomous multimodal path planning method, and a collaborative approach for aerial-ground robots will be the interest points for us to work on.









\bibliographystyle{ieeetr}
\bibliography{reference}

\end{document}